\documentclass{article}

\usepackage{spconf,amsmath,graphicx,hyperref}
\usepackage{cite}
\usepackage[capitalise]{cleveref}

\newcommand{\proposed}{TiCT}
\newcommand{\proposedfull}{\underline{T}ime-series \underline{i}n-\underline{C}ontext \underline{T}ransformer}

\usepackage{enumitem}
\setlist[itemize]{leftmargin=*, topsep=.0em, itemsep=0pt, parsep=0pt, partopsep=0pt}
\setlist[enumerate]{leftmargin=*, topsep=.0em, itemsep=0pt, parsep=0pt, partopsep=0pt}

\usepackage{collectbox}



\usepackage{multirow}

\newlength\savewidth
\newcommand\shline{\noalign{\global\savewidth\arrayrulewidth
                            \global\arrayrulewidth 1.5pt}%
                   \hline
                   \noalign{\global\arrayrulewidth\savewidth}
                   }

\usepackage{algorithm}
\usepackage[noend]{algpseudocode}

\newcommand*\Input[1]{\Statex \textbf{Input:} #1}
\newcommand*\Output[1]{\Statex \textbf{Output:} #1}
\algrenewcommand\alglinenumber[1]{#1}

\title{\proposed{}: A SYNTHETICALLY PRE-TRAINED FOUNDATION MODEL\\FOR TIME SERIES CLASSIFICATION}

\name{
    \begin{tabular}{@{}c@{}}
        Chin-Chia Michael Yeh \qquad Uday Singh Saini \qquad Junpeng Wang \qquad Xin Dai \\
        Xiran Fan \qquad Jiarui Sun \qquad Yujie Fan \qquad Yan Zheng
    \end{tabular}
}
\address{Visa Research}

\begin{document}
\ninept

\maketitle

\begin{abstract}
The ubiquity of time series data creates a strong demand for general-purpose foundation models, yet developing them for classification remains a significant challenge, largely due to the high cost of labeled data.
Foundation models capable of in-context learning (ICL) offer a powerful solution, adapting to new tasks with minimal examples and reducing the need for extensive retraining.
However, prior work on large-scale time series models has predominantly focused on forecasting, leaving a critical gap for versatile, fine-tuning-free classification.
To address this, we introduce \proposed{} (\proposedfull{}), a transformer-based model pre-trained exclusively on synthetic data to perform in-context classification.
We make two primary technical contributions: 1) a novel architecture featuring a scalable bit-based label encoding and a special output attention mechanism to handle an arbitrary number of classes; and 2) a synthetic pre-training framework that combines a Mixup-inspired process with data augmentation to foster generalization and noise invariance.
Extensive evaluations on the UCR Archive show that \proposed{} achieves competitive performance against state-of-the-art supervised methods.
Crucially, this is accomplished using only in-context examples at inference time, without updating a single model weight.
The source code is available at: \url{https://sites.google.com/view/tsicl}.
\end{abstract}



\section{Introduction}
\begin{figure}[htp]
\centerline{
\includegraphics[width=0.99\linewidth]{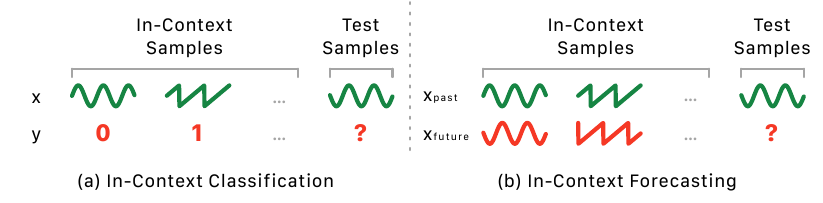}
}
\vspace{-1em}
\caption{
a) ICL for Classification, the focus of this work, predicts the class of a query series by conditioning on a context of labeled examples. 
b) ICL for Forecasting, the focus of most prior work, uses a historical context to predict future time series values.
}
\label{fig:learn_prob}
\vspace{-2em}
\end{figure}

Time series data is ubiquitous, generated by any system that produces measurable, time-varying signals across domains like medicine, human motion, and transportation~\cite{dau2019ucr}.
Developing specialized models for each new classification task, however, often requires extensive labeled data, which can be costly and time-consuming to acquire.
Foundation models capable of in-context learning (ICL) present a powerful alternative, as they can adapt to new, unseen tasks using only a handful of labeled examples provided as context.
Despite this promise, the majority of large-scale time series models have focused strictly on forecasting~\cite{ansari2024chronos,lu2024context,hoo2025tables,taga2025timepfn}, leaving the equally important problem of classification largely unaddressed, a distinction we illustrate in \cref{fig:learn_prob}.

To fill this gap, we introduce \proposed{} (\proposedfull{}), a foundation model designed specifically for ICL time series classification.
Our approach is centered on two core principles: a Transformer-based architecture to leverage its proven ICL capabilities~\cite{vaswani2017attention,garg2022can}, and an exclusive reliance on pre-training with synthetic data.
Pre-training on synthetic data is advantageous for two key reasons: 1) it overcomes the data scarcity limitations of public time series archives, and 2) it has been validated as a successful strategy for ICL models in adjacent domains like tabular data and time series forecasting~\cite{hollmann2022tabpfn,taga2025timepfn}.

However, designing such a model presents two distinct challenges that this work systematically addresses.
First, existing synthetically-trained classifiers like tabPFN~\cite{hollmann2022tabpfn,thomas2024retrieval,hollmann2025accurate} are often limited to a small, fixed number of classes (e.g., 10), which is insufficient for diverse benchmarks like the UCR Archive, where class counts range up to 60~\cite{dau2019ucr}.
To solve this, we introduce a scalable bit-based label representation and a special output attention mechanism that directly generates a probability for each potential class, allowing \proposed{} to handle an arbitrary number of classes.
Second, the pre-training process must create a diverse set of synthetic tasks that teach the model how to generalize.
Our strategy achieves this in two ways: we use a Mixup-inspired~\cite{zhang2017mixup,wickstrom2022mixing} process to generate coherent pairs of synthetic time series and labels, and simultaneously apply data augmentations---such as warping, shifting, and Gaussian noise---to ensure the model learns invariance to common time series distortions~\cite{wen2020time,iwana2021empirical,yeh2023toward}.

Through these contributions, \proposed{} achieves competitive performance on the UCR Archive datasets when compared to state-of-the-art supervised methods.
Notably, it accomplishes this entirely through ICL, without requiring any weight updates or task-specific fine-tuning.
In summary, our main contributions are:
\begin{itemize}
    \item A novel architecture for ICL classification featuring a scalable bit-based label encoding and a special output attention mechanism to generate class probabilities, a design with potential applications beyond time series.
    \item A synthetic pre-training framework that uses a Mixup-inspired process to generate labeled tasks and incorporates data augmentation to build invariance to common time series noise.
    \item A comprehensive evaluation showing that \proposed{} is an effective, fine-tuning-free foundation model for time series classification.
\end{itemize}
\section{Background}
\label{sec:background}
\noindent \textbf{In-Context Learning.}
A classification model exhibits in-context learning (ICL) capability if it can predict the class label $y_\text{test}$ of a previously unseen sample $x_\text{test}$ by conditioning on a context of labeled examples, $\mathcal{D}_\text{train}=\{(x_i, y_i)\}_i^n$~\cite{hollmann2022tabpfn,hollmann2025accurate}.
This is achieved by modeling the posterior predictive distribution $p_\theta(y_\text{test}|x_\text{test}, \mathcal{D}_\text{train})$, which is commonly formulated as a softmax over a function $f_\theta$ that scores potential labels:
{\footnotesize
\begin{equation}
    p_\theta(y_\text{test}|x_\text{test}, \mathcal{D}_\text{train})=\frac{\exp{f_\theta(x_\text{test}, \mathcal{D}_\text{train})[y_\text{test}]}}{\sum_{c=1}^C \exp{f_\theta(x_\text{test}, \mathcal{D}_\text{train})[c]}}
    \label{eq:icl_dist}
\end{equation}}

\noindent where $C$ is the number of classes in the context $\mathcal{D}_\text{train}$, and $[\cdot]$ denotes vector indexing.
In this work, we focus on a prominent variant known as retrieval-based ICL, where the context is not the entire training set but a dynamically retrieved subset of relevant examples~\cite{lewis2020retrieval,thomas2024retrieval}.
Specifically, the global context $\mathcal{D}_\text{train}$ is replaced by the set of $k$ nearest neighbors of the test sample $x_\text{test}$, denoted as $k\text{NN}(x_\text{test},\mathcal{D}_\text{train})$.
Consequently, \cref{eq:icl_dist} is modified to condition on this local context:
{\footnotesize
\begin{equation}
    p_\theta(y_\text{test}|x_\text{test}, \mathcal{D}_\text{train})=\frac{\exp{f_\theta(x_\text{test}, k\text{NN}(x_\text{test},\mathcal{D}_\text{train}))[y_\text{test}]}}{\sum_{c=1}^C \exp{f_\theta(x_\text{test}, k\text{NN}(x_\text{test},\mathcal{D}_\text{train}))[c]}}
\end{equation}}

\noindent where $k\text{NN}(\cdot)$ is a function that returns the $k$ nearest neighbors of $x_\text{test}$ from $\mathcal{D}_\text{train}$.
Our adoption of retrieval-based ICL is motivated by three primary advantages:
\begin{enumerate}
    \item \textit{Improved Expressivity:} By focusing on a local context of nearest neighbors, the model can capture intricate local patterns and avoids the potential underfitting that can occur when conditioning on a large, heterogeneous global context~\cite{thomas2024retrieval}.
    \item \textit{Enhanced Scalability:} The computational cost is significantly reduced, as the model only processes a small set of $k$ neighbors rather than the entire training set, making the approach scalable to large datasets~\cite{thomas2024retrieval}.
    \item \textit{Suitability for Time Series:} Nearest-neighbor algorithms are effective for time series~\cite{yeh2016matrix,yeh2017matrix,bagnall2017great,yeh2022error,yeh2024matrix}, which provides a strong prior for using a $k\text{NN}$-based retrieval mechanism in this domain.
\end{enumerate}

\noindent \textbf{Time Series Encoders.}
A crucial component of the model is the time series encoder, which transforms a raw time series input into a fixed-dimensional embedding for the ICL framework.
Based on their proven success in supervised time series classification, two prominent architectures are considered in this work: Residual Networks (ResNet) and Transformers.

\begin{enumerate}
    \item \textit{ResNet:} Residual Networks are a leading architecture for time series classification, often serving as a strong baseline in empirical studies~\cite{wang2017time,ismail2019deep,yeh2023toward}.
    Adapted for time series, the architecture typically consists of a stack of residual blocks built upon 1D convolutional layers, which are effective at capturing local patterns. 
    For our implementation, we adopt the specific design from~\cite{yeh2023toward}.
    \item \textit{Transformer:} The Transformer has been widely adopted for time series modeling due to its powerful attention mechanism~\cite{vaswani2017attention,yeh2023toward,ansari2024chronos}.
    When used as an encoder, it treats a time series as a sequence of tokens or patches, and its self-attention mechanism is highly effective at capturing long-range dependencies. 
    For our implementation, we adopt the specific design from~\cite{yeh2023toward}.
\end{enumerate}
\section{Methodology}
This section details the two core components of our proposed method, \proposed{}.
First, in \cref{sec:icl_model}, we describe the model's architecture, focusing on our novel approach to label representation and the special output attention mechanism that enables classification.
Second, in \cref{sec:pre_train}, we detail the synthetic data generation process and the overall pre-training algorithm that empowers the model's in-context learning capabilities.

\subsection{In-Context Learning Model Architecture}
\label{sec:icl_model}
\begin{figure}[htp]
\vspace{-1em}
\centerline{
\includegraphics[width=0.55\linewidth]{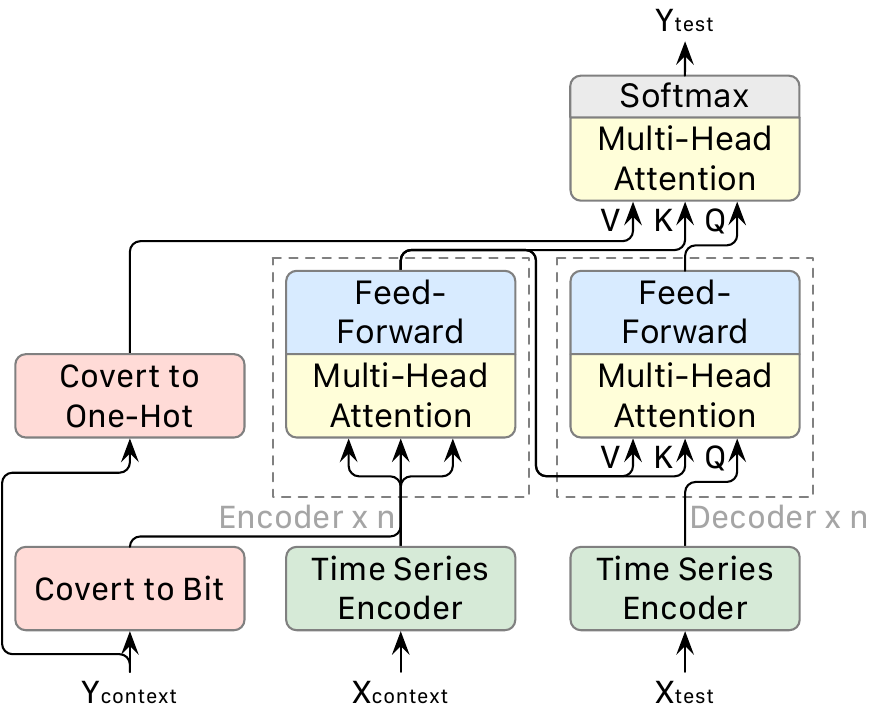}
}
\vspace{-1em}
\caption{
An overview of the \proposed{} architecture. The model processes in-context samples and labels through an encoder and uses a decoder to relate them to the test sample. A final output attention mechanism computes class probabilities by attending to the in-context labels.
}
\label{fig:icl_model}
\end{figure}

The architecture of \proposed{} is a custom Transformer-based model designed specifically for in-context classification, as illustrated in \cref{fig:icl_model}.
The model accepts three primary inputs: a set of in-context time series samples $X_\text{context}=\{x_i\}_i^k$, their corresponding labels $Y_\text{context}=\{y_i\}_i^k$, and a test sample $X_\text{test}$ to be classified.
The core of the model consists of a time series encoder, a Transformer encoder-decoder stack, and a special output attention mechanism, which we detail below.

\noindent \textbf{1. Input Representation and Encoding.}
All input time series ($X_\text{context}$ and $X_\text{test}$) are first transformed into fixed-dimensional embeddings, $E_\text{context}$ and $E_\text{test}$, using a shared time series encoder, as described in \cref{sec:background}.
The in-context labels, $Y_\text{context}$, are encoded into two distinct formats for different purposes.
First, a compact \textit{bit representation} of each label is concatenated with its corresponding time series embedding, $E_\text{context}$.
This combined representation is then fed into the Transformer encoder, allowing the model to learn associations between time series patterns and their symbolic class assignments.
Second, a \textit{one-hot representation} of the labels, $Y_\text{one-hot}$, is prepared for use in the final output attention layer.

\noindent \textbf{Justification for Bit-Based Label Encoding.}
For a model to generalize across in-context learning tasks, it cannot learn any intrinsic meaning associated with a specific label index (e.g., `0' or `1').
Instead, it must learn to treat labels as abstract, interchangeable symbols whose crucial information is the \textit{relationships} they define—specifically, which input samples belong to the same group~\cite{garg2022can, boix2023can}.
This requires the Transformer to learn a function that is \textit{equivariant} with respect to permutations of these symbolic labels.
We chose bit representation for the encoder input over alternatives like numerical or one-hot encoding for both theoretical and empirical reasons.
Compared to the numerical representation used in tabPFN-based models~\cite{hollmann2022tabpfn,hollmann2025accurate,thomas2024retrieval}, a bit representation enforces a ``discrete" symbolic nature, discouraging the model from interpreting labels as continuous values.
Compared to one-hot encoding, which can be sparse and high-dimensional (a vector of size $C$), the bit representation (a vector of size $\lceil\log_2 C\rceil$) is significantly more compact.
This denser representation simplifies the learning task, as each embedding dimension associated with a bit has a higher probability of being active (0.5) compared to the sparse activation ($\frac{1}{C}$) in a one-hot vector, assuming a uniform random class distribution.
As confirmed by our label reasoning experiments in \cref{sec:label_exp}, while a Transformer can be trained to treat labels as abstract symbols using any of the three representations, the bit representation converges fastest by a large margin.

\noindent \textbf{2. Transformer Encoder-Decoder Processing.}
The sequence of context embeddings combined with their label bits is processed by the Transformer encoder to produce context representations $H_\text{context}$.
The test embedding, $E_\text{test}$, and the encoder's output $H_\text{context}$ are then fed into the Transformer decoder, which produces a final test representation, $H_\text{test}$.
Crucially, the decoder in \proposed{} is modified to exclude the self-attention layer.
This design choice is deliberate: it ensures that the prediction for each test sample is made independently, without being influenced by other test samples that may be processed in the same batch.

\noindent \textbf{3. Output Attention and Label Equivariance.}
The final classification is performed by a special output attention mechanism that computes a probability distribution over the set of classes seen in the context.
This mechanism takes the final test representation $H_\text{test}$ as its query ($Q$), the context representations from the encoder $H_\text{context}$ as its keys ($K$), and the one-hot encoded context labels $Y_\text{one-hot}$ as its values ($V$).
The attention weights, $\alpha = \text{Softmax}\left( QK^T/\sqrt{d_k} \right)$, therefore represent the learned similarity between the test sample and each in-context sample.
The output logits, $l$, are computed as the weighted sum of the one-hot value vectors:
$ l = \sum_{i=1}^{k} \alpha_i V_i = \sum_{i=1}^{k} \alpha_i Y_{\text{one-hot}, i} $.
This formulation has a powerful and desirable property: for any class $c$ present in the in-context examples, its corresponding logit $l[c]$ is precisely the sum of the attention weights of all samples belonging to that class ($l[c] = \sum_{i \in \mathcal{I}_c} \alpha_i$).

This design ensures that the model's output is equivariant with respect to permutations of the class labels.
We now formally demonstrate this property. Let $\pi: \{1, \dots, C\} \to \{1, \dots, C\}$ be a permutation function on the class indices.
Assuming the Transformer encoder has been successfully trained to treat labels as abstract symbols, the query vector $Q$ and key vectors $K_i$ that it produces are equivariant.
Consequently, the attention weights $\alpha_i$, which measure the semantic similarity between the query and keys, remain invariant under this label permutation.
The one-hot value vectors $V_i$, however, are permuted directly.
Let $V_i$ be the original one-hot vector for a sample with label $c$, and let $V'_i$ be the new one-hot vector for the same sample, now with label $\pi(c)$.
The new output logits, denoted $l'$, are computed using the invariant weights $\alpha_i$ and the permuted values $V'_i$:
$ l' = \sum_{i=1}^{k} \alpha_i V'_i $.
The logit for a new class index $j$ is given by the sum of weights for all samples now labeled $j$, which means $y'_i = j$.
Since $y'_i = \pi(y_i)$, the condition $y'_i = j$ is equivalent to $y_i = \pi^{-1}(j)$.
Thus, the new logit for class $j$ is:
$ l'[j] = \sum_{i \text{ s.t. } y_i = \pi^{-1}(j)} \alpha_i $.
This is precisely the original logit for class $\pi^{-1}(j)$.
Therefore, we have $l'[j] = l[\pi^{-1}(j)]$.
This result shows that the new logits vector is a permutation of the original logits vector, determined by $\pi$.
This property formally demonstrates that the model does not learn the meaning of any specific class index, but instead learns a general classification mechanism based on the relationships within the in-context examples.

\subsection{Pre-training Framework}
\label{sec:pre_train}
The pre-training framework for \proposed{} is designed to teach the model how to perform in-context classification by exposing it to a vast number of diverse, synthetically generated classification tasks.
The core of this framework is a synthetic data generation algorithm, detailed in \cref{alg:pre_train}, which is inspired by the principles of Mixup~\cite{zhang2017mixup,wickstrom2022mixing} and incorporates a suite of time series data augmentations~\cite{wen2020time,iwana2021empirical,yeh2023toward}.
This process generates entire ``in-context datasets," each containing samples and corresponding labels in both bit and one-hot formats.

\begin{algorithm}[htp]
    \centering
    \caption{Synthetic In-Context Dataset Generation\label{alg:pre_train}}
    \footnotesize
    \begin{algorithmic}[1]
        \Input{dataset size~$n$, label bits~$n_\text{bit}$, max augmentations~$n_\text{max}$}
        \Output{samples~$X$, one-hot labels~$Y_\text{one-hot}$, bit labels~$Y_\text{bit}$}
        \Function{GenerateDataset}{$n, n_\text{bit}, n_\text{max}$}
            \State $\text{template}_0, \text{template}_1 \gets \textsc{KernelSynth()}, \textsc{KernelSynth()}$
            \State $(\text{one\_hot}_0, \text{bit}_0) \gets ([1,0], \textsc{RandBinary}(n_\text{bit}))$
            \State $(\text{one\_hot}_1, \text{bit}_1) \gets ([0,1], \textsc{RandBinary}(n_\text{bit})) \text{ s.t. } \text{bit}_1 \neq \text{bit}_0$
            \State $t \gets \mathcal{U}(0.1, 0.9)$
            \State $\textsc{AugF} \gets $ \Call{SelectRandAugs}{$n_\text{max}$}
            \For{$i = 1 \textbf{ to } n$}
                \State $r \gets \mathcal{U}(0, 1)$
                \State $x \gets r \cdot \text{template}_0 + (1-r) \cdot \text{template}_1$
                \State $x \gets \textsc{AugF}(x)$
                \State $(y_\text{one-hot}, y_\text{bit}) \gets \begin{cases} (\text{one\_hot}_0, \text{bit}_0) & \text{if } r > t \\ (\text{one\_hot}_1, \text{bit}_1) & \text{otherwise} \end{cases}$
                \State Append $(x, y_\text{one-hot}, y_\text{bit})$ to outputs
            \EndFor
            \State \Return $X, Y_\text{one-hot}, Y_\text{bit}$
        \EndFunction
    \end{algorithmic}
\end{algorithm}

\noindent \textbf{Synthetic Task Generation.}
The generation process, outlined in \cref{alg:pre_train}, begins by creating a simple binary classification problem.
Two distinct time series shape templates are generated using KernelSynth~\cite{ansari2024chronos}, a method well-suited for creating realistic and diverse time series (Line 4).
We intentionally use only two templates to construct a single, clear decision boundary for each synthetic task.
The underlying hypothesis is that if a model can learn to discriminate between two classes, this fundamental capability can generalize to multi-class problems, analogous to how SVMs are extended to multi-class settings.
We validate this empirically in \cref{sec:ucr_exp}.
Correspondingly, labels are created for these two classes (Lines 5--6).
The one-hot labels are fixed to $[1,0]$ and $[0,1]$, but the bit-based labels are generated as unique, random binary vectors.
Training on these randomized bit vectors forces the model to treat them as abstract symbols, enabling it to generalize to tasks with up to $2^{n_\text{bit}}$ classes.

\noindent \textbf{Promoting Generalization and Robustness.}
To enhance the model's generalization, we introduce two sources of randomness.
First, a class imbalance threshold $t$ is sampled from $\mathcal{U}(0.1, 0.9)$ for each dataset to simulate real-world scenarios where class distributions may be skewed (Line 7).
Second, a pipeline is created by randomly selecting between zero and $n_\text{max}$ time series augmentations\footnote{Jittering, smoothing, slope, spike, step, cropping, masking, shifting, and time warping~\cite{yeh2023toward}.} (Line 8).
Inside the generation loop (Lines 9--13), each new sample $x$ is created by first mixing the two templates with a random ratio $r$ (Line 10).
This sample is then passed through the augmentation pipeline (Line 11).
While the set of augmentations is the same for all samples within a synthetic dataset, the stochastic nature of each function (e.g., the location of a spike) ensures instance-level diversity.
Finally, a class label is assigned based on whether the mixing ratio $r$ exceeds the threshold $t$ (Line 12).

\noindent \textbf{Pre-training Procedure.}
The pre-training process treats each dataset generated by \cref{alg:pre_train} as a single data point for the ICL model.
For each generated dataset, we split the $n$ samples in half.
The first half serves as the in-context examples ($X_\text{context}, Y_\text{context}$), while the second half serves as the query samples ($X_\text{test}$) and their corresponding ground-truth labels ($Y_\text{test}$).
To form a training batch of size $n_\text{batch}$, we invoke \cref{alg:pre_train} $n_\text{batch}$ times, creating a batch of distinct in-context learning tasks.
The model is then trained end-to-end using a standard cross-entropy loss between its predicted probabilities and the ground-truth labels $Y_\text{test}$.
\section{Experiments}
We validate our design choices and evaluate the performance of \proposed{} through two primary experiments: an ablation study on label representation strategies in \cref{sec:label_exp}, and a benchmark against baselines on the UCR Archive in \cref{sec:ucr_exp}~\cite{dau2019ucr}.

\subsection{Study on Label Representation}
\label{sec:label_exp}
A core hypothesis of our work is that the choice of label representation significantly impacts a Transformer's ability to learn the symbolic reasoning required for ICL.
To test this, we designed a synthetic task to isolate and measure how effectively a Transformer learns to perform label matching using three different encoding strategies: numerical, one-hot, and our proposed bit representation.

\noindent \textbf{Experimental Setup.}
The task is a binary matching problem.
For each data point, we generate an in-context set with labels $Y=\{y_i\}_i^n$.
These labels are drawn from a universe of $2^{n_\text{bit}}$ possible classes.
A query class label, $q$, is selected, and each label $y_i$ in the context is set to either $q$ or another randomly chosen label, $p \neq q$.
The model is then trained to output `1' for any context item $i$ where $y_i=q$ and `0' otherwise.
Essentially, the model must learn to identify which context items match the query label, a fundamental capability for ICL classification.

We trained three separate Transformer models, one for each label representation strategy.
All models share the same core architecture (8 layers, 128 embedding dimensions, 4 attention heads) and training hyperparameters.
The only architectural difference is the input layer that maps the label representation to the 128-dimensional embedding space: a $1 \to 128$ linear layer for the numerical representation, a $2^{n_\text{bit}} \to 128$ layer for one-hot, and an $n_\text{bit} \to 128$ layer for our bit representation.
We set the context size $n=15$ and the number of bits $n_\text{bit}=8$.
The models were trained for 400 epochs, with each epoch consisting of 4096 randomly generated in-context datasets.
Performance was measured by accuracy on a fixed test set of 1024 pre-generated datasets.

\begin{figure}[htp]
\vspace{-1em}
\centerline{
\includegraphics[width=0.9\linewidth]{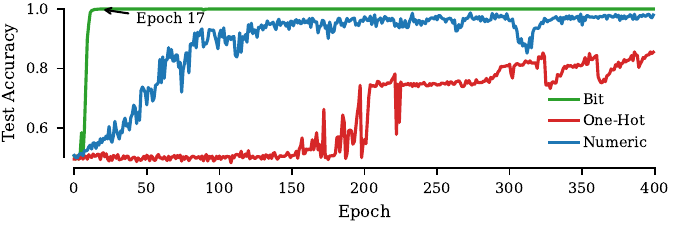}
}
\vspace{-1em}
\caption{
Test accuracy over 400 training epochs for three label representation strategies. The bit representation demonstrates significantly faster convergence and superior final performance, reaching perfect accuracy by the 17th epoch.
}
\label{fig:comp_rep}
\end{figure}

\noindent \textbf{Results and Analysis.}
The learning curves for the three strategies are shown in \cref{fig:comp_rep}.
The results provide strong empirical support for our design choice.
While all three representations eventually show signs of learning, the model using our proposed \textit{bit representation }converges dramatically faster and reaches near-perfect test accuracy by only the 17th epoch.
This superior learning efficiency validates our hypothesis that a compact and dense representation simplifies the task of learning symbolic label relationships, a key finding that informed the final design of \proposed{}.

\subsection{Evaluation on the UCR Archive}
\label{sec:ucr_exp}
We evaluate \proposed{} on all 128 datasets from the UCR Time Series Archive~\cite{dau2019ucr}, a standard benchmark for time series classification.

\noindent \textbf{Evaluation Protocol.}
We use the official train/test splits for each of the 128 datasets. 
In-context examples for each test sample are retrieved from the corresponding training set using z-normalized Euclidean distance, with a context size of $k=64$. 
We report classification accuracy and use average rank across all datasets for aggregated comparison, where a lower rank is better.

\noindent \textbf{Ablation Study.}
We first conduct an ablation study to validate our design choices, with results in \cref{tab:ablation}. 
The findings confirm that each component of our pre-training framework is crucial. 
Removing data augmentation (- Augmentation) severely degrades performance (rank 5.01 vs. 3.13), highlighting its role in building noise invariance. 
Replacing our binary Mixup strategy with a multi-class approach (- Mixup) also worsens the rank to 4.25.
This result suggests that pre-training on the continuum of examples generated between two base classes via Mixup is a more effective generalization strategy than training on discrete multi-class templates.
Disabling the class imbalance simulation (- Imbalance) similarly harms performance (rank 3.39). 
The study also confirms ResNet as the superior encoder for \proposed{} and demonstrates that scaling the model size (\proposed{}-Large) substantially boosts performance, achieving the top average rank of 2.04.

\begin{table}[htp]
\vspace{-1em}
\caption{
Ablation study of \proposed{} on the UCR Archive. 
We report the average rank (lower is better). 
}
\label{tab:ablation}
\begin{center}
\resizebox{0.99\linewidth}{!}{%
\begin{tabular}{l|cccc|c}
\shline
\textbf{Model Variant} & \textbf{Encoder} & \textbf{Mixup} & \textbf{Augment.} & \textbf{Imbalance} & \textbf{Avg. Rank} \\ \hline \hline
\proposed{} - Augmentation & ResNet & Yes & No & Yes & 5.01 \\
\proposed{} - Mixup & ResNet & No & Yes & Yes & 4.25 \\
\proposed{} - Imbalance & ResNet & Yes & Yes & No & 3.39 \\ \hline
\proposed{} (Transformer) & Transformer & Yes & Yes & Yes & 3.17 \\
\proposed{} (ResNet, 2M) & ResNet & Yes & Yes & Yes & 3.13 \\
\proposed{}-Large (ResNet, 47M) & ResNet & Yes & Yes & Yes & \textbf{2.04} \\
\shline
\end{tabular}%
}
\end{center}
\vspace{-1em}
\end{table}

\noindent \textbf{Comparison with Baselines.}
We compare \proposed{} against established baselines in \cref{tab:performance}, including classic methods (1NN) and fully-supervised deep learning models (ResNet, Transformer).
Crucially, the supervised baselines require training 128 separate models—one for each dataset.
In contrast, \proposed{} uses a single, pre-trained foundation model for all 128 datasets with no fine-tuning.
Remarkably, our \proposed{}-Large model (Avg. Rank 2.86) performs on par with the fully-supervised ResNet (2.86) and is competitive with the supervised Transformer (2.58).
A Wilcoxon signed-rank test confirms that these performance differences are not statistically significant.
This result is very significant, showing that a single, fine-tuning-free foundation model can match the performance of task-specific supervised models, validating our approach's practical value.

\begin{table}[htp]
\vspace{-1em}
\caption{
Comparison of \proposed{} against baselines on all 128 UCR Archive datasets. 
\proposed{} is a single foundation model, while ResNet and Transformer are fully supervised (128 trained models each).
}
\label{tab:performance}
\begin{center}
\resizebox{0.8\linewidth}{!}{%
\begin{tabular}{l|cc|cc|c}
\shline
 & \multicolumn{2}{c|}{Classic} & \multicolumn{2}{c|}{Supervised Deep Learning} & ICL \\
& 1NN-ED & 1NN-DTW & ResNet & Transformer & \proposed{} \\ \hline
Avg. Rank & 3.55 & 3.15 & 2.86 & 2.58 & 2.86 \\
\shline
\end{tabular}%
}
\end{center}
\vspace{-1em}
\end{table}

\vspace{-1em}

\section{Conclusion}
In this work, we introduced \proposed{}, a foundation model that fills a critical gap in in-context learning for time series classification. 
Its architecture uses a scalable bit-based label representation and a special output attention mechanism to handle a variable number of classes, while a Mixup-inspired pre-training strategy builds invariance to common distortions. 
Our comprehensive evaluation shows that \proposed{} achieves competitive performance on the UCR Archive through in-context learning alone, without any weight updates.
Future work will address the interaction between time series and large language models~\cite{yeh2025empowering} and extend our in-context learning approach to the financial domain~\cite{yeh2025treasure}.

\bibliographystyle{IEEEbib}
{\footnotesize
\bibliography{section/reference}

@inproceedings{yeh2023toward,
  title={Toward a foundation model for time series data},
  author={Yeh, Chin-Chia Michael and Dai, Xin and Chen, Huiyuan and Zheng, Yan and Fan, Yujie and Der, Audrey and Lai, Vivian and Zhuang, Zhongfang and Wang, Junpeng and Wang, Liang and others},
  booktitle={Proceedings of the 32nd ACM International Conference on Information and Knowledge Management},
  pages={4400--4404},
  year={2023}
}

@article{dau2019ucr,
  title={The UCR time series archive},
  author={Dau, Hoang Anh and Bagnall, Anthony and Kamgar, Kaveh and Yeh, Chin-Chia Michael and Zhu, Yan and Gharghabi, Shaghayegh and Ratanamahatana, Chotirat Ann and Keogh, Eamonn},
  journal={IEEE/CAA Journal of Automatica Sinica},
  volume={6},
  number={6},
  pages={1293--1305},
  year={2019},
  publisher={IEEE}
}

@article{ansari2024chronos,
  title={Chronos: Learning the language of time series},
  author={Ansari, Abdul Fatir and Stella, Lorenzo and Turkmen, Caner and Zhang, Xiyuan and Mercado, Pedro and Shen, Huibin and Shchur, Oleksandr and Rangapuram, Syama Sundar and Arango, Sebastian Pineda and Kapoor, Shubham and others},
  journal={arXiv preprint arXiv:2403.07815},
  year={2024}
}

@article{hoo2025tables,
  title={From Tables to Time: How TabPFN-v2 Outperforms Specialized Time Series Forecasting Models},
  author={Hoo, Shi Bin and M{\"u}ller, Samuel and Salinas, David and Hutter, Frank},
  journal={arXiv preprint arXiv:2501.02945},
  year={2025}
}

@article{lu2024context,
  title={In-context time series predictor},
  author={Lu, Jiecheng and Sun, Yan and Yang, Shihao},
  journal={arXiv preprint arXiv:2405.14982},
  year={2024}
}

@inproceedings{taga2025timepfn,
  title={TimePFN: Effective multivariate time series forecasting with synthetic data},
  author={Taga, Ege Onur and Ildiz, Muhammed Emrullah and Oymak, Samet},
  booktitle={Proceedings of the AAAI Conference on Artificial Intelligence},
  volume={39},
  number={19},
  pages={20761--20769},
  year={2025}
}

@article{thomas2024retrieval,
  title={Retrieval \& fine-tuning for in-context tabular models},
  author={Thomas, Valentin and Ma, Junwei and Hosseinzadeh, Rasa and Golestan, Keyvan and Yu, Guangwei and Volkovs, Maks and Caterini, Anthony L},
  journal={Advances in Neural Information Processing Systems},
  volume={37},
  pages={108439--108467},
  year={2024}
}

@article{hollmann2025accurate,
  title={Accurate predictions on small data with a tabular foundation model},
  author={Hollmann, Noah and M{\"u}ller, Samuel and Purucker, Lennart and Krishnakumar, Arjun and K{\"o}rfer, Max and Hoo, Shi Bin and Schirrmeister, Robin Tibor and Hutter, Frank},
  journal={Nature},
  volume={637},
  number={8045},
  pages={319--326},
  year={2025},
  publisher={Nature Publishing Group UK London}
}

@article{hollmann2022tabpfn,
  title={Tabpfn: A transformer that solves small tabular classification problems in a second},
  author={Hollmann, Noah and M{\"u}ller, Samuel and Eggensperger, Katharina and Hutter, Frank},
  journal={arXiv preprint arXiv:2207.01848},
  year={2022}
}

@article{vaswani2017attention,
  title={Attention is all you need},
  author={Vaswani, Ashish and Shazeer, Noam and Parmar, Niki and Uszkoreit, Jakob and Jones, Llion and Gomez, Aidan N and Kaiser, {\L}ukasz and Polosukhin, Illia},
  journal={Advances in neural information processing systems},
  volume={30},
  year={2017}
}

@article{garg2022can,
  title={What can transformers learn in-context? a case study of simple function classes},
  author={Garg, Shivam and Tsipras, Dimitris and Liang, Percy S and Valiant, Gregory},
  journal={Advances in neural information processing systems},
  volume={35},
  pages={30583--30598},
  year={2022}
}

@article{wen2020time,
  title={Time series data augmentation for deep learning: A survey},
  author={Wen, Qingsong and Sun, Liang and Yang, Fan and Song, Xiaomin and Gao, Jingkun and Wang, Xue and Xu, Huan},
  journal={arXiv preprint arXiv:2002.12478},
  year={2020}
}

@article{iwana2021empirical,
  title={An empirical survey of data augmentation for time series classification with neural networks},
  author={Iwana, Brian Kenji and Uchida, Seiichi},
  journal={Plos one},
  volume={16},
  number={7},
  pages={e0254841},
  year={2021},
  publisher={Public Library of Science San Francisco, CA USA}
}

@article{zhang2017mixup,
  title={mixup: Beyond empirical risk minimization},
  author={Zhang, Hongyi and Cisse, Moustapha and Dauphin, Yann N and Lopez-Paz, David},
  journal={arXiv preprint arXiv:1710.09412},
  year={2017}
}

@article{ismail2019deep,
  title={Deep learning for time series classification: a review},
  author={Ismail Fawaz, Hassan and Forestier, Germain and Weber, Jonathan and Idoumghar, Lhassane and Muller, Pierre-Alain},
  journal={Data mining and knowledge discovery},
  volume={33},
  number={4},
  pages={917--963},
  year={2019},
  publisher={Springer}
}

@inproceedings{wang2017time,
  title={Time series classification from scratch with deep neural networks: A strong baseline},
  author={Wang, Zhiguang and Yan, Weizhong and Oates, Tim},
  booktitle={2017 International joint conference on neural networks (IJCNN)},
  pages={1578--1585},
  year={2017},
  organization={IEEE}
}

@inproceedings{yeh2016matrix,
  title={Matrix profile I: all pairs similarity joins for time series: a unifying view that includes motifs, discords and shapelets},
  author={Yeh, Chin-Chia Michael and Zhu, Yan and Ulanova, Liudmila and Begum, Nurjahan and Ding, Yifei and Dau, Hoang Anh and Silva, Diego Furtado and Mueen, Abdullah and Keogh, Eamonn},
  booktitle={2016 IEEE 16th international conference on data mining (ICDM)},
  pages={1317--1322},
  year={2016},
  organization={IEEE}
}

@inproceedings{yeh2017matrix,
  title={Matrix profile VI: Meaningful multidimensional motif discovery},
  author={Yeh, Chin-Chia Michael and Kavantzas, Nickolas and Keogh, Eamonn},
  booktitle={2017 IEEE international conference on data mining (ICDM)},
  pages={565--574},
  year={2017},
  organization={IEEE}
}

@inproceedings{yeh2022error,
  title={Error-bounded approximate time series joins using compact dictionary representations of time series},
  author={Yeh, Chin-Chia Michael and Zheng, Yan and Wang, Junpeng and Chen, Huiyuan and Zhuang, Zhongfang and Zhang, Wei and Keogh, Eamonn},
  booktitle={Proceedings of the 2022 SIAM International Conference on Data Mining (SDM)},
  pages={181--189},
  year={2022},
  organization={SIAM}
}

@inproceedings{yeh2024matrix,
  title={Matrix Profile for Anomaly Detection on Multidimensional Time Series},
  author={Yeh, Chin-Chia Michael and Der, Audrey and Saini, Uday Singh and Lai, Vivian and Zheng, Yan and Wang, Junpeng and Dai, Xin and Zhuang, Zhongfang and Fan, Yujie and Chen, Huiyuan and others},
  booktitle={2024 IEEE International Conference on Data Mining (ICDM)},
  pages={911--916},
  year={2024},
  organization={IEEE}
}

@article{bagnall2017great,
  title={The great time series classification bake off: a review and experimental evaluation of recent algorithmic advances},
  author={Bagnall, Anthony and Lines, Jason and Bostrom, Aaron and Large, James and Keogh, Eamonn},
  journal={Data mining and knowledge discovery},
  volume={31},
  number={3},
  pages={606--660},
  year={2017},
  publisher={Springer}
}

@article{wickstrom2022mixing,
  title={Mixing up contrastive learning: Self-supervised representation learning for time series},
  author={Wickstr{\o}m, Kristoffer and Kampffmeyer, Michael and Mikalsen, Karl {\O}yvind and Jenssen, Robert},
  journal={Pattern Recognition Letters},
  volume={155},
  pages={54--61},
  year={2022},
  publisher={Elsevier}
}

@article{lewis2020retrieval,
  title={Retrieval-augmented generation for knowledge-intensive nlp tasks},
  author={Lewis, Patrick and Perez, Ethan and Piktus, Aleksandra and Petroni, Fabio and Karpukhin, Vladimir and Goyal, Naman and K{\"u}ttler, Heinrich and Lewis, Mike and Yih, Wen-tau and Rockt{\"a}schel, Tim and others},
  journal={Advances in neural information processing systems},
  volume={33},
  pages={9459--9474},
  year={2020}
}

@article{boix2023can,
  title={When can transformers reason with abstract symbols?},
  author={Boix-Adsera, Enric and Saremi, Omid and Abbe, Emmanuel and Bengio, Samy and Littwin, Etai and Susskind, Joshua},
  journal={arXiv preprint arXiv:2310.09753},
  year={2023}
}

@article{yeh2025empowering,
  title={Empowering Time Series Forecasting with LLM-Agents},
  author={Yeh, Chin-Chia Michael and Lai, Vivian and Saini, Uday Singh and Fan, Xiran and Fan, Yujie and Wang, Junpeng and Dai, Xin and Zheng, Yan},
  journal={arXiv preprint arXiv:2508.04231},
  year={2025}
}

@article{yeh2025treasure,
  title={{TREASURE}: A Transformer-Based Foundation Model for
High-Volume Transaction Understanding},
  author={Yeh, Chin-Chia Michael and Saini, Uday Singh and Dai, Xin and Fan, Xiran and Jain, Shubham and Fan, Yujie and Sun, Jiarui and Wang, Junpeng and Pan, Menghai and Dou, Yingtong Chen, Yuzhong and Rakesh, Vineeth and Wang, Liang and Zheng, Yan and Das, Mahashweta },
  journal={arXiv preprint arXiv:2511.19693},
  year={2025}
}
}

\end{document}